\documentclass{article} 
\usepackage{jmlr2e-arxiv}
\usepackage{graphicx,subfigure}
\usepackage[T1]{fontenc}
\usepackage[latin9]{inputenc}
\usepackage{amsmath}
\usepackage{amssymb}
\usepackage{multirow}
\usepackage{esint}
\usepackage{color}
\usepackage{natbib}
\usepackage{url}

\usepackage{algorithm}
\usepackage{algorithmic}
\usepackage{hyperref}

\begin{document} 
\newcommand{\beq}{\begin{equation}}
\newcommand{\eeq}{\end{equation}}
\newcommand{\beqa}{\begin{eqnarray}}
\newcommand{\eeqa}{\end{eqnarray}}
\newcommand{\beqas}{\begin{eqnarray*}}
\newcommand{\eeqas}{\end{eqnarray*}}

\newcommand{\bit}{\begin{itemize}}
\newcommand{\eit}{\end{itemize}}
\newcommand{\bits}{\begin{itemize*}}
\newcommand{\eits}{\end{itemize*}}
\newcommand{\benum}{\begin{enumerate}}
\newcommand{\eenum}{\end{enumerate}}
\newcommand{\benums}{\begin{enumerate*}}
\newcommand{\eenums}{\end{enumerate*}}
\newcommand{\comment}[1]{}


\newcommand{\selfco}{{\tt GC}}
\newcommand{\selfcomin}{{\tt GC}$^{-1}$}


\newcommand{\epps}{\epsilon}
\newcommand{\eppsh}{\hat{\epsilon}}
\newcommand{\lam}{\lambda}
\newcommand{\dis}{{D}}
\newcommand{\M}{\mathcal{M}}
\newcommand{\T}{\mathcal{T}}
\newcommand{\D}{\mathcal{D}}
\newcommand{\dataset}{\D}
\newcommand{\nsamp}{N}
\newcommand{\dhigh}{r}
\newcommand{\Lbw}{L}
\newcommand{\bw}{\epsilon} 
\newcommand{\mbx}{x}  

\newcommand{\RR}{\mathbb{R}} 
\newcommand{\1}{\mathbf{1}}
\newcommand{\indicator}[1]{\mathbbm{1}_{\left[ {#1} \right] }}
\newcommand{\rf}{\mathbf{r}}

\newcommand{\fracpartial}[2]{\frac{\partial {#1}}{\partial {#2}}}
\newcommand{\bigOO}{\mathcal{O}}
\newcommand{\diag}[1]{{\rm diag}\{\,#1\}}



\newcommand{\qed}{\nobreak \ifvmode \relax \else%
      \ifdim\lastskip<1.5em \hskip-\lastskip
      \hskip1.5em plus0em minus0.5em \fi \nobreak
      \vrule height0.75em width0.5em depth0.25em\fi}

\newcommand{\lyxdot}{.}


\newcommand{\phase}{\theta}
\newcommand{\X}{\mathcal{X}}
\newcommand{\G}{\mathcal{G}}

\newcommand{\dM}{\text{dim}(\M)}
\newcommand{\N}{{\mathcal N}}
\newcommand{\Ln}{\tilde{\mathcal{L}}_{\epps_\nsamp,\nsamp}}

\newcommand{\Lbwnot}{\mathcal{L}_{\epps,\nsamp}} 
\newcommand{\LB}{\Delta_{\M}} 
\newcommand{\Lop}{\mathcal{L}}
\newcommand{\Ho}{\mathcal{H}}
\newcommand{\Dstar}{\D^{*}}

\newcommand{\mby}{y} 
\newcommand{\mbz}{z} %
\newcommand{\mbxstar}{\mbx^*}
\newcommand{\density}{\pi}

\newcommand{\dembed}{s}
\newcommand{\dintri}{d}  

\newcommand{\ihigh}{g_{\mbox{\raisebox{-0.2ex}{\tiny $\!\RR^\dhigh$}}}}  
\newcommand{\iembed}{\delta_{\dembed}}  
\newcommand{\inc}{\jmath} 


\newcommand{\matrg}{{\mathbf G}}
\newcommand{\matrh}{{\mathbf H}}
\newcommand{\matri}{{\mathbf I}}
\newcommand{\matrj}{{\mathbf J}}
\newcommand{\matrl}{{\mathbf L}}
\newcommand{\matrp}{{\mathbf P}}
\newcommand{\matrs}{{\mathbf S}}
\newcommand{\matrt}{{\mathbf T}}
\newcommand{\matrx}{{\mathbf X}}

\newcommand{\matrlap}{\matrl}

\newcommand{\myprod}{\cdot} 


\newlength{\picwi}  

\newcommand{\fix}{\marginpar{FIX}}
\newcommand{\new}{\marginpar{NEW}}
\newcommand{\theHalgorithm}{\arabic{algorithm}}

\title{Improved Graph Laplacian via Geometric Self-Consistency}

\author{Dominique Perrault-Joncas\thanks{Amazon.com, Seattle, USA, email:\texttt{joncas@amazon.com}}, %
Marina Meil\u{a}\thanks{University of Washington, Seattle, USA, email:\texttt{mmp@stat.washington.edu}}}

\maketitle

\begin{abstract}
  We address the problem of setting the kernel bandwidth $\epps$ used
  by Manifold Learning algorithms to construct the graph Laplacian.
  Exploiting the connection between manifold geometry, represented by
  the Riemannian metric, and the Laplace-Beltrami operator, we set
  $\epps$ by optimizing the Laplacian's ability to preserve the
  geometry of the data. Experiments show that this principled approach
  is effective and robust.
\end{abstract}

\section{Introduction}

Manifold learning and manifold regularization are popular tools for
dimensionality reduction and clustering \cite{BelNiy02,LuxBelBou08}, as
well as for semi-supervised
learning \cite{BelNiy06,Zhu03,ZhouBelkin11,SmoKon03} and
modeling with Gaussian Processes \cite{sind07}. Whatever the task,
a manifold learning method requires the user to provide an external
parameter, called ``bandwidth'' or ``scale'' $\epps$, that defines the size of the local
neighborhood.

More formally put, a common challenge in semi-supervised and
unsupervised manifold learning lies in obtaining a ``good'' graph
Laplacian estimator $\Lbw$. We focus on the practical problem of
optimizing the parameters used to construct $\Lbw$ and, in particular,
$\epps$. As we see empirically, since the Laplace-Beltrami operator on
a manifold is intimately related to the geometry of the manifold, our
estimator for $\epps$ has advantages even in methods that do not
explicitly depend on $\Lbw$.

In manifold learning, there has been sustained interest for
determining the asymptotic properties of $\Lbw$
\cite{GinKol06,belniy07,HeiAudLux07,TingHJ10}.  The most relevant is
\cite{Singer06}, which derives the optimal rate for $\epps$ w.r.t. the
sample size $\nsamp$ \beq \epps^2 =
C(\M)\nsamp^{-\frac{1}{3+\dintri/2}} \\, \label{eq:singer} \eeq with
$\dintri$ denoting the intrinsic dimension of the data manifold
$\M$. The problem is that $C(\M)$ is a constant that depends on the
yet unknown data manifold, so it is rarely known in practice. Also, this result is asymptotic, in the limit of very large sample sizes. 

Considerably fewer studies have focused on the
parameters used to construct $\Lbw$ in a finite sample problem. A
common approach is to ``tune'' parameters by cross-validation in the semi-supervised context. However, in an unsurpervised problem like
non-linear dimensionality reduction, there is no context in which to
apply cross-validation. While several approaches \cite{Lee07NDR,ChenBuja:localMDS09,Levb,Bias} may yield a usable parameter, they generally do not aim to improve $\Lbw$
\textit{per se} and offer no geometry-based justification for its selection. 

In this paper, we present a new, geometrically inspired approach to
selecting the bandwidth parameter $\epps$ of $\Lbw$ for a given data
set. It is known that, in the hypothesis that the data come from a
manifold $\M$, the {\em Laplace-Beltrami} operator $\Delta_\M$ on the
data manifold $\M$ contains all the intrinsic geometry of $\M$. Hence,
we compare the geometry induced by the graph Laplacian $\Lbw$ with the local
data geometry and choose the value of $\epps$ for which these two are closest.

\section{Background: Heat Kernel, Laplacian and Geometry}

Our paper builds on two previous sets of results: 1) the construction of $\Lbw$ that is consistent for $\Delta_\M$ when the
sample size $\nsamp\rightarrow\infty$ under the manifold hypothesis (see \cite{CoiLaf06}); and 2) the
relationship between $\Delta_\M$ and the Riemannian metric $g$ on a manifold, as well as the estimation of $g$ (see \cite{JoncasMeila12}).

{\bf Construction of the graph Laplacian.} Several methods could be used to construct $\Lbw$ (see \cite{HeiAudLux07,TingHJ10}). The one we present, due to \cite{CoiLaf06}, guarantees that,
if the data are sampled from a manifold $\M$, $\Lbw$ converges to
$\Delta_\M$:

Given a set of points $\dataset=\{x_1,\dots,x_\nsamp\}$ in
high-dimensional Euclidean space $\RR^\dhigh$, construct a
weighted graph ${\cal G}=(\dataset,W)$ over them, with
$W=[W_{ij}]_{ij=1:N}$. The weight $W_{ij}$ between $x_i$ and $x_j$ is
the {\em heat kernel}
\cite{BelNiy02}
\beq
W_{ij}\,\equiv\,W_{\epsilon}(x_{i},x_{j})\, = \, \exp\left(\left|\left|x_{i}-x_{j}\right|\right|_{2}^{2}/\epsilon^2\right),
\label{eq:heat_kernel}
\eeq
with $\epps$ a {\em bandwidth} parameter fixed by the user. Next, construct $\Lbw=[\Lbw_{ij}]_{ij}$ of ${\cal G}$ by
\beqa     
\!\!\!\!d_i &\!\!\!\!\!\! =&\!\!\!\!\!\! \sum_j W_{ij}\,,
\;\;\;
W'_{ij} = \frac{W_{ij}}{d_id_j}\,,
\;\;\;
d'_i = \sum_j W'_{ij}
\,,\;
\text{and} \, \,  
\Lbw_{ij} = \sum_j \frac{W'_{ij}}{d'_j}\,.
      \label{eq:cont_L} 
\eeqa 
Equation \eqref{eq:cont_L} represents the discrete
versions of the renormalized Laplacian construction from \cite{CoiLaf06}.
Note that $d_i,d'_i,W',\Lbw$ all depend on the bandwidth $\epps$ via
the heat kernel. 


{\bf Estimation of the Riemannian metric.} We follow
\cite{JoncasMeila12} in this step. A {\em Riemannian manifold}
$(\M,g)$ is a {\em smooth manifold} $\M$ endowed with a {\em
Riemannian metric} $g$; the metric $g$ at point $p\in \M$ is a scalar
product over the vectors in $\T_p\M$, the {\em tangent subspace} of
$\M$ in $p$. In any coordinate representation of $\M$, $g_p\equiv
G(p)$ -- the Riemannian metric at $p$ -- represents a positive definite
matrix\footnote{This paper contains mathematical objects like $\M$,
$g$ and $\Delta$, and computable objects like a data point $x$, and the
graph Laplacian $\Lbw$. The Riemannian metric {\em at a point} belongs to
both categories, so it will sometimes be denoted $g_p,g_{x_i}$ and
sometimes $G(p),G(x_i)$, depending on whether we refer to its mathematical
or algorithmic aspects. This also holds for the dual metric $h$, defined in Proposition \ref{prop:l-g}.} of dimension $d$ equal to the {\em
intrinsic dimension} of $\M$. The significance of the metric $g$ as a repository of the geometry of $\M$ arises mainly from two facts: (i)
the {\em volume element} for any integration over $\M$ is given by
$\sqrt{\det G(x)} dx$, and (ii) the {\em line element} for computing
distances along a curve $x(t)\subset
\M$ is $\sqrt{\left(\frac{dx}{dt}\!\right)\!^T\!G(x)\frac{dx}{dt}}$. 

If we assume that the data we observe (in $\RR^\dhigh$) lies on a
manifold, then in the original coordinates, the metric $G(p)$ is the
unit matrix of dimension $\dintri$ padded with zeros up to dimension
$\dhigh$. When the data is mapped to another coordinate system -- for
instance by a manifold learning algorithm that performs non-linear
dimension reduction -- the matrix $G(p)$ changes with the coordinates
to reflect the distortion induced by the mapping (see \cite{JoncasMeila12} for more details). 

\begin{proposition}\label{prop:l-g}
Let $x$ denote local coordinate functions of a smooth Riemannian manifold $(\M,g)$ of dimension
$d$ and $\Delta_\M$ the Laplace-Beltrami operator defined on $\M$.
Then:\\
1.\cite{Ros97}  For any function $f\in {\cal C}^2(\M)$ 
\[
\Delta_\M f\;=\;\frac{1}{\sqrt{\text{det}(G)}}
\sum_{l=1}^d\frac{\partial}{\partial x^l}\left(\sqrt{\text{det}(G)}\sum_{k=1}^d
(G^{-1})_{lk}\frac{\partial}{\partial x^k}f\right).
\] 

2. $H(p)=(G(p))^{-1}$ the (matrix) inverse of the
Riemannian metric at point $p$, is given by
\beq \label{eq:g_inv} 
(H(p))^{ij}\;=\;\frac{1}{2}\Delta_{\M}\left(x^{i}-x^{i}(p)\right)\left(x^{j}-x^{j}(p)\right)|_{x=x(p)}
\eeq
with $i,j=1,\dots,d$
\end{proposition}
In \eqref{eq:g_inv} above, the right hand side is the application of
the $\Delta_\M$ operator to the function
$\left(x^{i}-x^{i}(p)\right)\left(x^{j}-x^{j}(p)\right)$, where
$x^i,x^j$ denote coordinates $i,j$ seen as functions on $\M$ and $x(p)$ 
is the coordinate map evaluated at point $p\in\M$. The
inverse matrices $(g_p)^{-1}=h_p\equiv H(p)$, being symmetric and
positive definite, determine a Riemannian metric $h$ called the {\em
dual metric} on $\M$.

Proposition \ref{prop:l-g} shows that the geometry of a smooth
manifold $\M$ is completely encoded by $\Delta_\M$ and, conversely, that $g$ completely
determines $\Delta_\M$. Through \eqref{eq:g_inv}, it also
provides a way to estimate $g$ from data.  Algorithm \ref{alg:rmetric}, adapted from \cite{JoncasMeila12}, implements
\eqref{eq:g_inv}. 

\begin{algorithm}[tb]
   \caption{Riemannian Metric($X,i,L,dual\in\{-1,1\}$)}
   \label{alg:rmetric}
\begin{algorithmic}
   \STATE {\bfseries Input:} $\nsamp \times \dintri$ design matrix $X$, $i$ index in data set, Laplacian $\Lbw$, binary variable $dual$
   \FOR{$k = 1 \to d$, $l = 1\to d$}
         \STATE $H_{k,l} \gets\sum_{j=1}^{\nsamp} L_{ij} 
\left ( X_{jk} -X_{ik})(X_{jl}- X_{il} \right )$ 
    \ENDFOR
	\RETURN $H^{dual}$  (i.e. $H$ if $dual=1$ and $H^{-1}$ if $dual=-1$)
\end{algorithmic}
\end{algorithm}

\section{A Quality Measure for $\Lbw$}

Having established that the Laplace-Beltrami operator on a manifold
$\M$ encodes the intrinsic geometry of $\M$, we propose to estimate
$\epsilon$ by optimizing how faithfully the corresponding $\Lbw$
captures the original data geometry. For this we must: (1) estimate the geometry $g$ both from $\Lbw$ and
without $\Lbw$ (Section \ref{sec:rmetric-robust}), and (2) define a
measure of agreement between the two (Section \ref{sec:distort}).

\subsection{\bf The Geometric Consistency Idea for Optimizing $\Lbw$}

We consider the trivial embedding of the data in the ambient space
$\RR^\dhigh$ for which the geometry is trivially known. This provides
a target $g$; we tune the scale of the Laplacian so that the $g$
calculated from Proposition \ref{prop:l-g} matches this target. Hence,
we choose $\epps$ to maximize {\em self-consistency} in the geometry
of the data.

More precisely, if $\M\subset\RR^\dhigh$ and inherits its metric from
$\RR^\dhigh$, as per the generally assumed hypothesis for
dimensionality reduction, then the Riemannian metric of $\mathcal{M}$
is $\ihigh|_{T\mathcal{M}}$.  Here, $\ihigh|_{T\mathcal{M}}$ stands
for the restriction of the natural metric of the ambient space
$\RR^\dhigh$ to the {\em tangent bundle} $T\mathcal{M}$ of the
manifold $\mathcal{M}$. We propose to tune the parameters of the graph
Laplacian $\Lbw$ so as to approximately enforce (a discrete,
coordinate expression of) the identity
\begin{equation} 
g_p\equiv \ihigh|_{T_{p}\mathcal{M}}\,\forall
p\in\mathcal{M}\,.\label{eq:equiv}
\end{equation}
In the above, the l.h.s. will be the metric implied from the Laplacian via Proposition \ref{prop:l-g}, and the r.h.s will be described below.
Mathematically speaking, \eqref{eq:equiv} is necessary and sufficient for finding the ``correct'' Laplacian. 


Note also that the geometric self-consistency approach is not limited to
the bandwidth parameter $\epps$, but can be applied to any other
parameter used in the construction of the Laplacian.

\subsection{Robust Estimation of the Metric}
\label{sec:rmetric-robust}
\comment{Equation \eqref{eq:equiv} represents our desired goal in terms of the
ideal mathematical objects of operators and manifold.} 

Exploiting equivalence \eqref{eq:equiv} to optimize the graph Laplacian
involves estimating $g$ from $\Lbw$ as prescribed by
Proposition \ref{prop:l-g} and representing the r.h.s $\ihigh|_{T_{p}\mathcal{M}}$numerically. Doing the latter directly via equation \eqref{eq:g_inv} is possilble, but naive, since it will yield a $\dhigh\times \dhigh$ matrix of rank $\dintri$\comment{This matrix would have rank equal to $d$ the intrinsic
dimension of $\M$ and its column/row range would span/be affinely
equivalent with the tangent subspace at $x$.}. Computing such a large
matrix is both inefficient and sensitive to noise in the data.


Instead, we estimate the tangent bundle $T\M$ and reduce the required
computations for \eqref{eq:g_inv} from $\nsamp\dhigh^{2}$ to $\nsamp
d^{2}$ by performing them directly on $T\M$. Specifically, we evaluate
the tangent subspace around each sampled point $x_i$ using local
Principal Component Analysis (PCA) and then express
$\ihigh|_{T_{p}\M}$ directly in the resulting low-dimensional subspace
as the unit matrix $I_d$. The tangent subspace also serves to define a
local coordinate chart, which is passed as input to Algorithm
\ref{alg:rmetric}, which computes $g_p$ in these coordinates.

When we compute $T_{x_i}\M$, for the sake of consistency, and to
ensure that the geometry we encode is common to all the
transformations we perform, we equate the notion of neighborhood in
the local PCA with that embodied in the heat kernel by choosing the
same bandwidth $\epps$ in both\footnote{In our experiments, we also
  implemented a version of our method that does not equate the two
  bandwidths. Since this did not yield improved performance, we have
  omitted it for brevity.}. This means that we conduct a {\em weighted
  local PCA (wlPCA)}, with weights defined by the heat kernel used to
produce the graph Laplacian \eqref{eq:heat_kernel}, centered around
$x_i$. This approach is similar to sample-wise weighted PCA of
\cite{Yue04}, with two important requirements: the weights must decay
rapidly away from $x_i$, and the data must be centered to have zero
mean such that all the points far from $x_i$ are mapped close to the
origin. These are satisfied by the weighted recentered design matrix
$Z$, where $Z_{j:}$, row $j$ of $Z$, is given by:
\beq \label{eq:z_recentered} Z_{j:}\,=\, \frac{W_{ij}
  (x_j-\bar{\mbx})}{\sum_{j'=1}^\nsamp W_{ij'}} \, ,\,\,\,
\text{with}\; \bar{\mbx}\,=\, \sum_{j=1}^\nsamp
\frac{W_{ij}x_j}{\sum_{j'=1}^\nsamp W_{ij'}}\,.  \eeq \cite{AsBiTo11}
proves that the wlPCA using the heat kernel, and equating the PCA and
heat kernel neighborhoods as we do, yields a consistent estimator of
$T_{x_i}\M$. This is implemented in Algorithm \ref{alg:tpproj}.

\begin{algorithm}[tb]
   \caption{Tangent Plane Projection($X,w,d$)    \label{alg:tpproj}}
\begin{algorithmic}
   \STATE {\bfseries Input:} $\nsamp \times \dhigh$ design matrix $X$, weight vector $w=[W_{i1}\,\ldots\,W_{iN}]$, dimension $d$
   \STATE Compute $Z,\,\bar{x}$ using~\eqref{eq:z_recentered}
   \STATE $[V,\Lambda] \gets \text{eig}(Z^t Z, d)$ ($d$-SVD of $Z$) 
   \STATE Center $X$ around  $\bar{x}$ from~\eqref{eq:z_recentered}
   \STATE $Y \gets X V_{:,1:d}$ (Project $X$ on $d$ principal subspace)
   \RETURN Y 
\end{algorithmic}
\end{algorithm}

In summary, to estimate the Riemannian metric at a point $x_i\in \D$,
one must (i) construct the graph Laplacian by \eqref{eq:cont_L}; (ii) perform
Algorithm \ref{alg:tpproj} to obtain $Y$; and (iii) apply Algorithm \ref{alg:rmetric} to $Y$ to obtain $G(\mbx_i)\in \RR^{d\times d}$. This
matrix is then compared with $I_d$.

We now take this approach a few steps further in terms of improving
its robustness with minimal sacrifice to its theoretical
grounding. First, it is debatable whether inverting $H$ in Algorithm
\ref{alg:rmetric} is necessary. Relation~\eqref{eq:equiv} is trivially
satisfied for the inverse Riemannian metric $\ihigh^{-1}$ since in the
chosen coordinates both $\ihigh$ and $\ihigh^{-1}$ are equal to the
unit matrix $I_d$. Therefore we will use the dual metric $h$ in place
of $g$ by default. Second, we perform both Algorithm \ref{alg:tpproj}
and Algorithm \ref{alg:rmetric} in $d'$ dimensions, with $d'\leq d$.

These changes make the algorithm faster, and make the computed dual metric $H$ both more stable
numerically and more robust to possible noise in the
data\footnote{We know from matrix perturbation theory that noise affects  the
$d$-th principal vector increasingly with $d$.}. Proposition \ref{prop:yvh'} shows that the resulting method remains theoretically sound.

\begin{proposition}\label{prop:yvh'} Let $X,\,Y,\,Z,\,V,\,W_{:i},\,H$, and $\dintri\geq 1$ represent the quantities in Algorithms \ref{alg:rmetric} and \ref{alg:tpproj}; assume that the columns of $V$ are sorted in decreasing order of the singular values, and that the rows and columns of $H$ are sorted according to the same order. Now denote by $Y',\,V',\,H'$ the quantitities computed by Algorithms \ref{alg:rmetric} and \ref{alg:tpproj} for the same $X,\,W_{:i}$ but with $d\leftarrow d'=1$. Then,
\beq \label{eq:yvh'}
V'\,=\,V_{:1}\in\RR^{\dhigh\times 1}
\;\;
Y'\,=\,Y_{:1}\in \RR^{\nsamp\times 1}
\;\;
H'\,=\,H_{11}\in \RR.
\eeq
\end{proposition}
The proof of this result is straightforward and omitted for
brevity. It is easy to see that Proposition \ref{prop:yvh'}
generalizes immediately to any $1\leq d'<d$. In other words, by using
$d'<d$, we will be projecting the data on a proper subspace of
$T_{x_i}\M$ -- namely, the subspace of least curvature \cite{Lee97}. The
dual metric $H'$ of this projection is the principal submatrix
of order $d'$ of $H$, i.e. $H_{11}$ if $d'=1$. Therefore, with the
reduced rank algorithms, we will only be enforcing a submatrix of $H$
to be close to the unit matrix. \comment{, leaving open the possibility that in
the orthogonal complement of $Y'$ in $T_{x_i}\M$ the metric has
eigenvalues smaller than 1. {\em REVIEW THE WE WILL BE ENFORCING PART}}  

\subsection{Measuring the Distortion}
\label{sec:distort}

For a finite sample, we cannot expect \eqref{eq:equiv} to hold
exactly, and so we need to define a distortion between the two metrics
to evaluate how well they agree.  We propose the {\em distortion}
\beq\label{eq:d-data} \dis\;=\;\frac{1}{N}\sum_{i=1}^N||H(x_i)-I_d||
\eeq where $||A||=\lambda_{max}(A)$ is the matrix spectral norm. Thus
$\dis$ measures the average distance of $H$ from the unit matrix over
the data set. For a ``good'' Laplacian, the distortion $\dis$ should
be minimal: \beq \hat{\epps}\;=\; \text{argmin}_{\epps}\dis
\,.\label{eq:argmin_d} \eeq
Before moving on, we note that the spectral norm in \eqref{eq:d-data}
is not chosed arbitrarily.  The expression of $\dis$ in
\eqref{eq:d-data} is the discrete version of the distance function
$D_{g_0}$ on the space of Riemannian metrics of a manifold $\M$
defined by
\beq \label{eq:distort}
\dis_{g_0}\left(g_1,g_2\right)\,=\,
 \int_{\M}\left|\left|g_1-g_2\right|\right|_{g_0}dV_{g_0},
\eeq
with volume element $dV_{g_0}=\sqrt{\text{det}G_0(x)}dx$ and 
\beq \label{eq:tensor}
\bigl.\left|\left|g\right|\right|_{g_0}\bigr|_p\;=\;\sup_{u,v\in \T_p\M\setminus\{0\}}
\frac{<u,v>_{g_p}}{<u,v>_{g_{0p}}}.
\eeq
Furthermore, the right-hand side of \eqref{eq:tensor} above represents
the {\em tensor norm} of $g_p$ on $\T_p\M$ with respect to the
Riemannian metric $g_{0p}$.  Now, \eqref{eq:d-data} follows when
$g_0,g_1,g_2$ are replaced by $I,\,I$ and $H$, respectively.




With \eqref{eq:argmin_d}, we have established a principled criterion
for selecting the parameter(s) of the graph Laplacian, by minimizing
the distortion between the true geometry and the geometry derived from
Proposition \ref{prop:l-g}. Practically, we compute $\dis$ by
Algorithm \ref{alg:distort} for each candidate $\epps$ , then choose
$\hat{\epps}$ by \eqref{eq:argmin_d}.

\begin{algorithm}[tb]
   \caption{Compute Distortion$(X,\epps,d)$}
   \label{alg:distort}
\begin{algorithmic}
   \STATE {\bfseries Input:} $\nsamp\times \dhigh$ design matrix $X$, $\epps$, working dimension $d$
   \STATE Compute the heat kernel $W$ by \eqref{eq:heat_kernel} for each pair of points in $X$
   \STATE Compute the graph Laplacian $L$ from $W$ by \eqref{eq:cont_L}
   \STATE $\dis \gets 0$
   \FOR{$i = 1 \to n$} 
        \STATE $Y \gets$ TangentPlaneProjection($X,W_{i,:},d$)
        \STATE $H \gets$ RiemannianMetric($Y,L,dual=1$) 
        \STATE $\dis \gets \dis+||H-I_d||^2/\nsamp$
   \ENDFOR 
   \RETURN $\dis$
\end{algorithmic}
\end{algorithm}

\comment{Computing $W$ and the graph Laplacian is $O(\nsamp^2)$, and is done
only once. Depending on the size of $\nsamp$ and $\dhigh$, one of the
most expensive steps is to compute the lwPCA in $\dhigh$ dimensions,
which is $O(\nsamp\dhigh^2)$ with a large constant. \comment{This is
the main disadvantage of using weighted local PCA rather than local
PCA based on $k-$nearest neighbors since $k+1$ points only span $k$
dimensions instead of $\dhigh$.} Algorithm \ref{alg:rmetric} takes
$O(\nsamp\dintri^2)$. 

For very high dimensional data or very large data sets, one can
significantly speed up all of the above computations by (approximative)
sparse matrix techiniques. These would turn any dependence on $\nsamp$
above into a dependence on $k$, where $k$ is the size of the largest
neighborhood.

A complementary possibility for reducing computing time is to compute
the distortion $\dis$ on a subsample, i.e. to select $\nsamp' \ll
\nsamp$ points at which to compute $\left | \left | H(x_i)-I \right |
\right| $, but still use all $\nsamp$ points in Algorithms
\ref{alg:rmetric} and \ref{alg:tpproj}. If the subsample is unbiased, this
will only affect the variance of the estimated distortion. We have
already mentioned that using a smaller value $d'$ for the working
dimension also speeds up computation.}

\section{Related Work}

Although the problem of estimating the ``scale'' of the data is
pervasive in manifold learning, work has focused mainly on asymptotic
results, with very few papers proposing estimation methods that can be
implemented in practice.

We have already mentioned the asymptotic result~\eqref{eq:singer} of
\cite{Singer06}. Other work in this area (\cite{GinKol06,HeiAudLux07,TingHJ10}) 
provides the necessary rates of change for $\epps$ with respect to
$\nsamp$ to guarantee convergence. These studies are relevant;
however, they all depend on manifold parameters that are usually not
known.\comment{ \textit{a priori}. Finding a self-consistent {\em why
self-consitent??} method to estimate these parameters in tandem with
estimating $\epps$ remains an open problem.}

Among practical methods, the most interesting is
that of \cite{ChenBuja:localMDS09}, which estimates $k$, the number of
nearest neighbors to use in the construction of the graph
Laplacian. It is reminiscent of our method, in that it is
self-consistent and evaluates a given $k$ with respect to the preservation of
$k'$ neighborhoods in the original
data. \comment{\cite{ChenBuja:localMDS09} offers a statistically
principled way to compare this criterion over different $k$ and $k'$
values.} However, it is not known how a method for estimating $k$ can be
translated into a method for estimating $\bw$ or vice versa (the two
graph construction methods exhibit different asymptotic behaviour
precisely because they give rise to different ensembles of
neighborhoods \cite{TingHJ10}).

Moreover, the method of \cite{ChenBuja:localMDS09} is designed to
optimize for a specific embedding, so the values obtained for $k$
depend on the embedding algorithm used. By contrast, the selection
algorithm we propose estimates an {\em intrinsic} quantity, a
scale $\bw$ that depends exclusively on the data. It is known \cite{GolZakKusRit08}
that most embeddings induce distortion in the data
geometry. Therefore, it is not clear that minimizing reconstruction
error for a particular method - Laplacian Eigenmap, for example - is optimal, since
even in the limit of infinite data, the embedding will distort the
original geometry.

Finally, we mention the algorithm proposed in
\cite{Guangliang_ChenLittleMaggioniRosasco:multiscale11} (CLMR). Its
goal is to obtain an estimate of the intrinsic dimension of the data;
however, a by-product of the algorithm is a range of scales where the
tangent space at a data point is well aligned with the principal
subspace obtained by a local singular value decomposition. As these are scales at which the
manifold looks locally linear, one can reasonably expect that they are
also the correct scales at which to approximate differential
operators, such as $\Delta_\M$. Given this, we implement the method
and compare it to our own results.

\section{Experimental Results}
{\bf Synthethic Data.} We experimented with estimating the bandwidth
$\eppsh$ on data sampled from known manifolds with
noise. We considered the two-dimensional {\tt hourglass} and {\tt
dome} manifolds of Figure \ref{fig:synth-epps}. We sampled uniformly from these manifolds, adding 10 ``noise'' dimensions and Gaussian noise ${\cal
N}(0,\sigma^2)$ to the resulting 13 dimensions.

The range of $\epps$ values was delimited by $\epps_{min}$ and
$\epps_{max}$. We set $\epps_{max}$ to the average of $||x_i-x_j||^2$
over all point pairs\comment{, a scale that is surely larger than the
local neighborhood for any manifold with $n>>d$. The lower limit} and
$\epps_{min}$ to the limit in which the heat kernel $W$ becomes
approximately equal to the unit matrix; this is tested by
$\max_j(\sum_iW_{ij})-1<\gamma$\footnote{Guaranteeing that all
eigenvalues of $W$ are less than $\gamma$ away from 1.} for $\gamma\approx
10^{-4}$. This range spans about two orders of magnitude in the data
we considered, and was searched by a logarithmic grid with
approximately 20 points. We saved computatation time by evaluating all pointwise quantities
($\hat{\dis}$, local SVD) on a random sample of size $\nsamp'=200$ of each
data set. 
We replicated each experiment on 10 independent samples.

\setlength{\picwi}{0.28\textwidth}
\begin{figure*}
\hskip 0.5cm
\begin{tabular}{cccc}
& ${\mathsf \sigma=0.001}$ 
& \hspace{-2.3em}${\mathsf \sigma=0.01}$ 
& \hspace{-2.3em}${\mathsf \sigma=0.1}$\\
\includegraphics[width=.9\picwi]{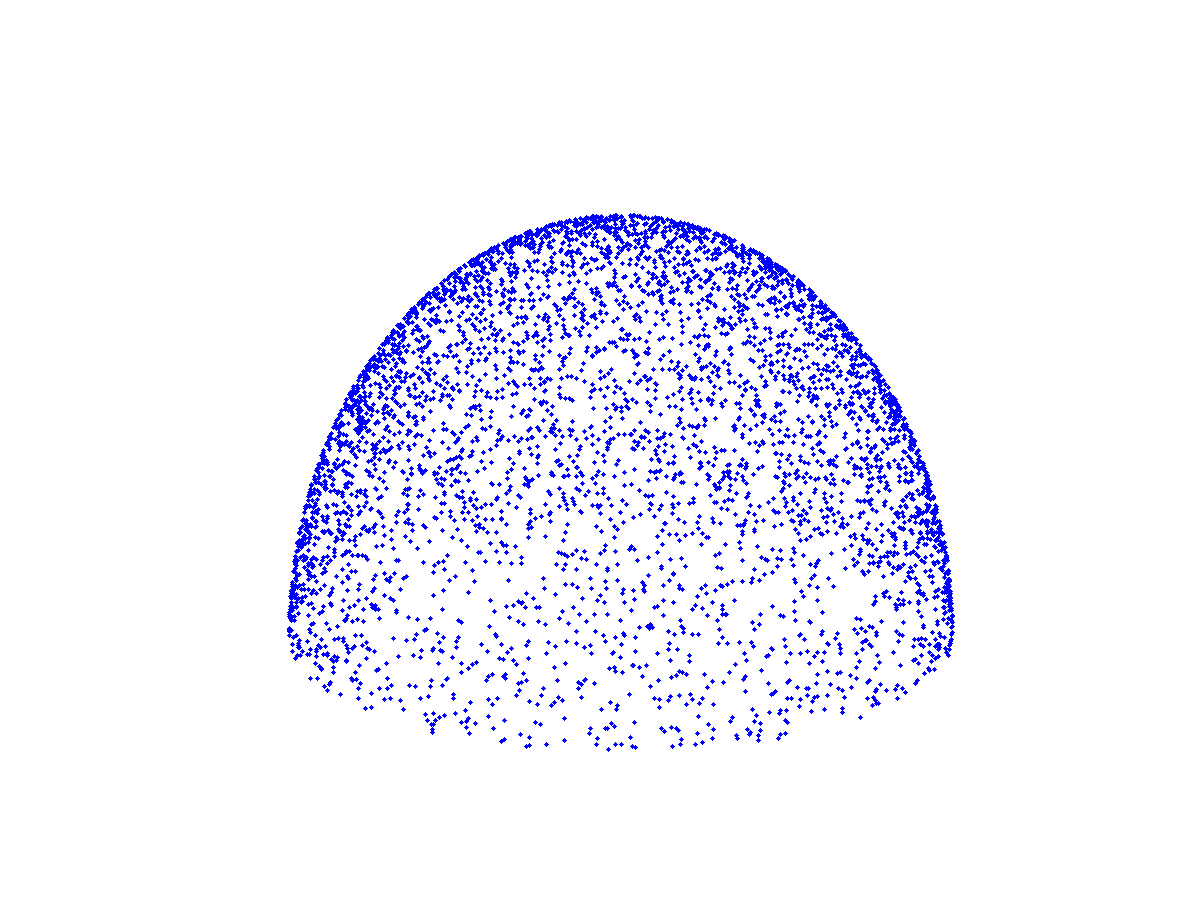}&
\hspace{-2.3em}
\includegraphics[width=\picwi]{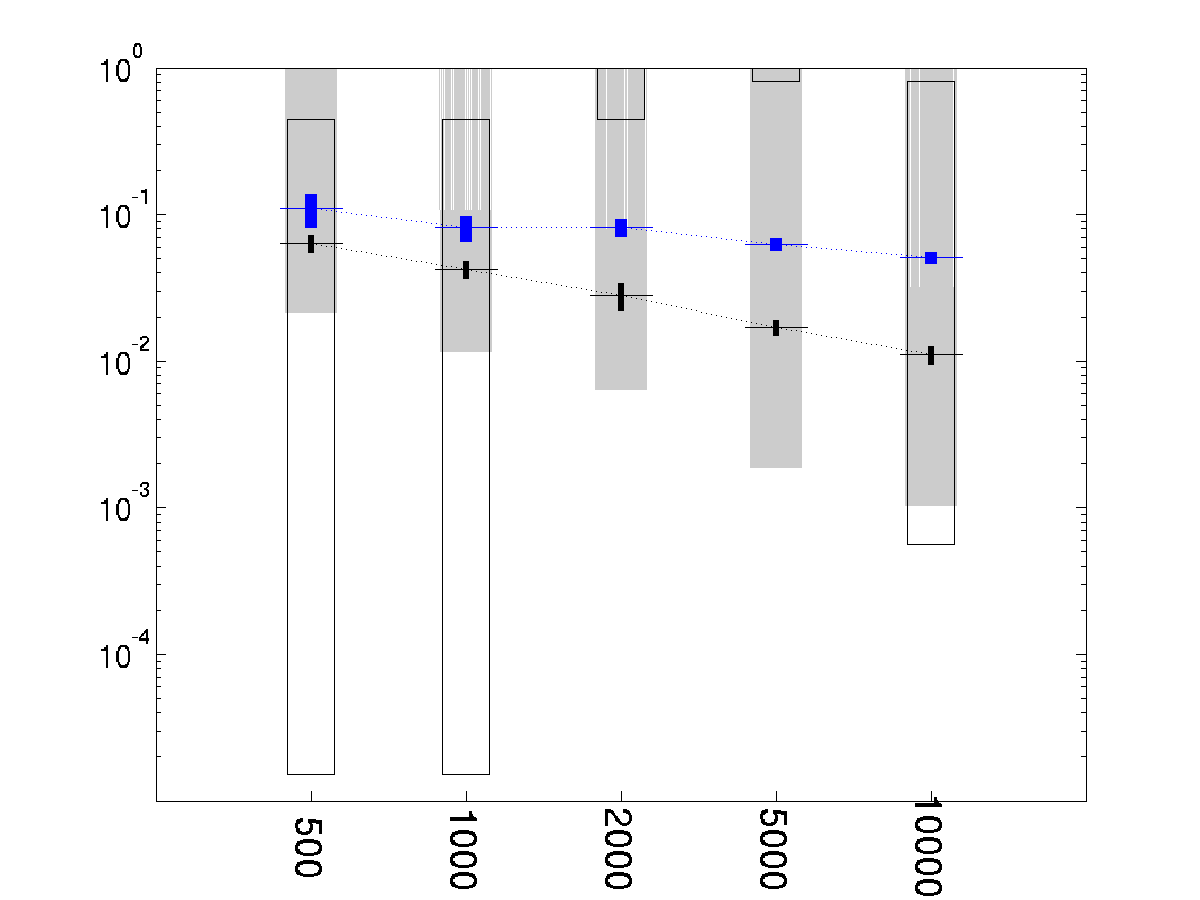}
&
\hspace{-2.3em}
\includegraphics[width=\picwi]{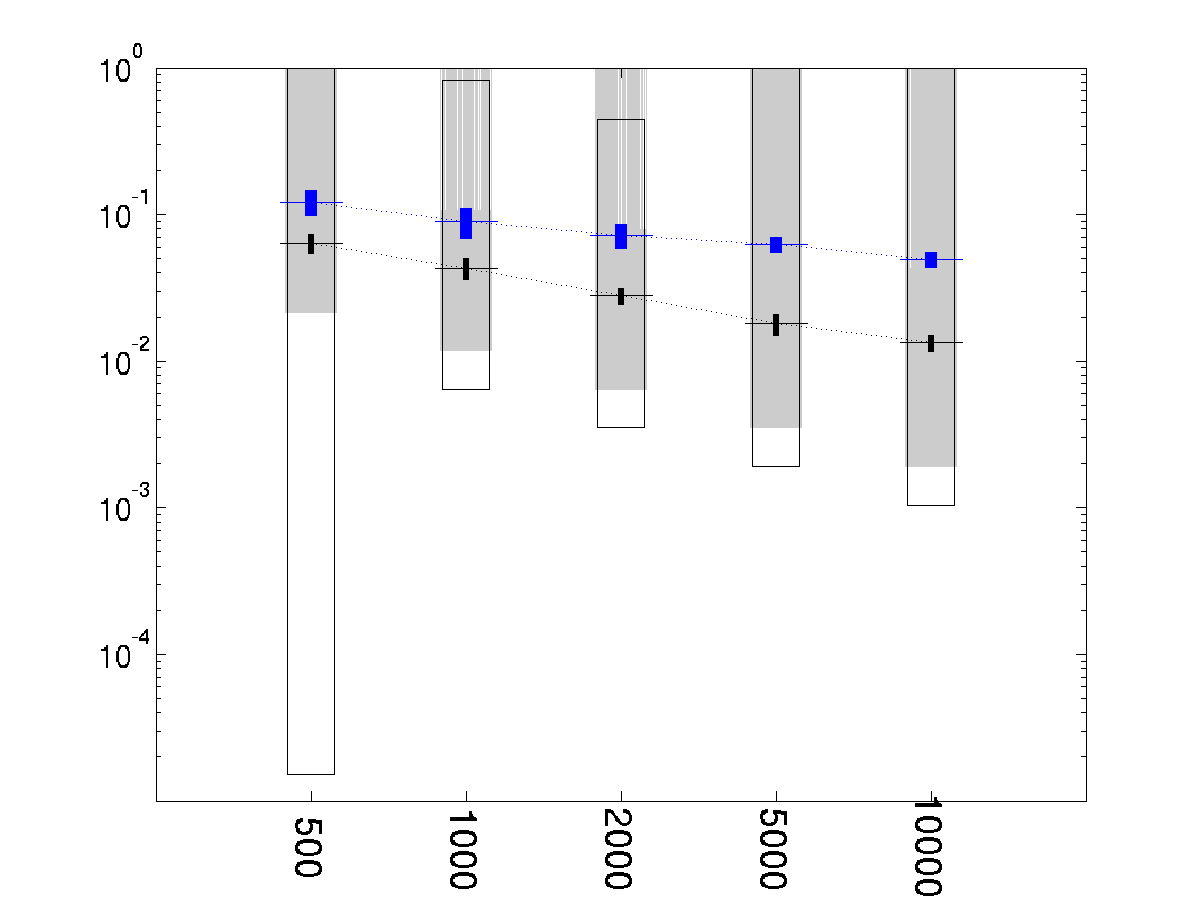}
&
\hspace{-2.3em}
\includegraphics[width=\picwi]{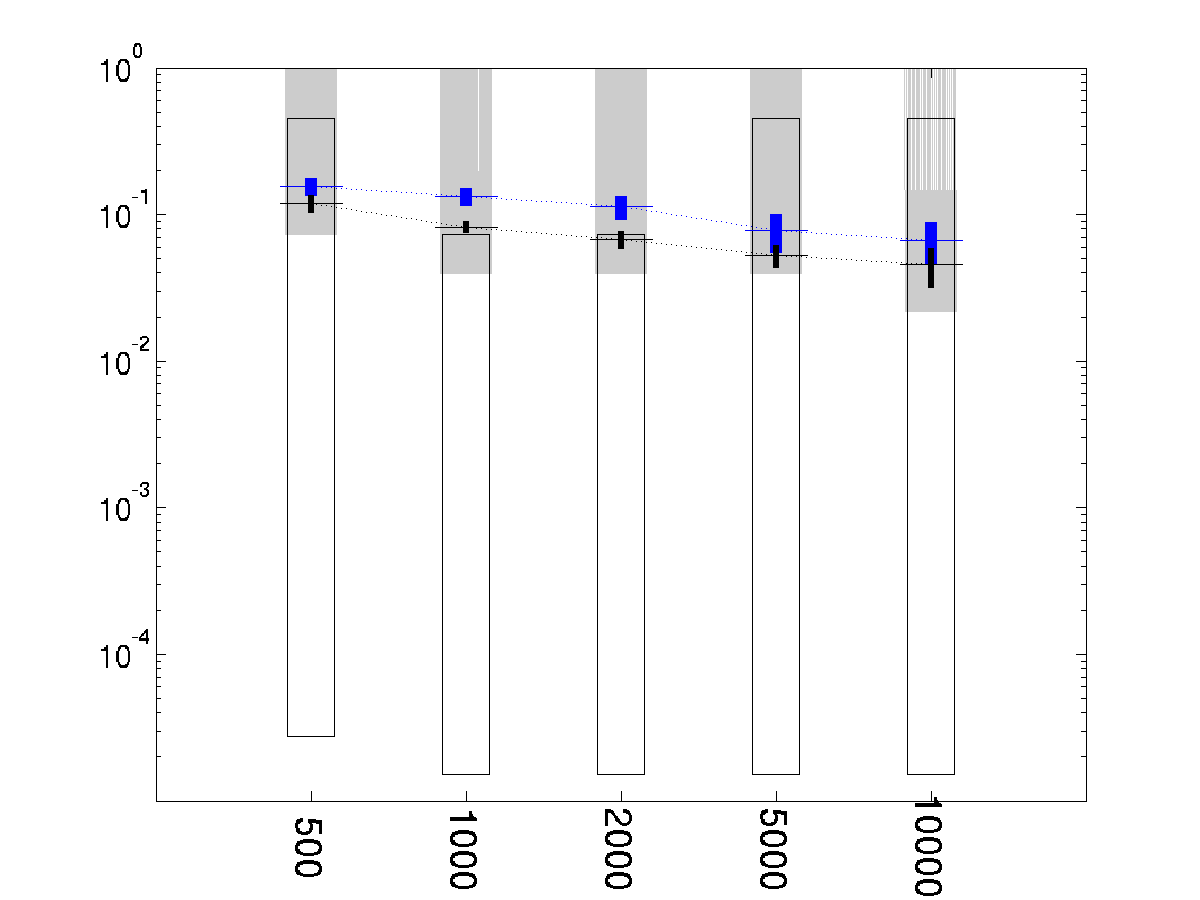}
\\
\includegraphics[width=1\picwi]{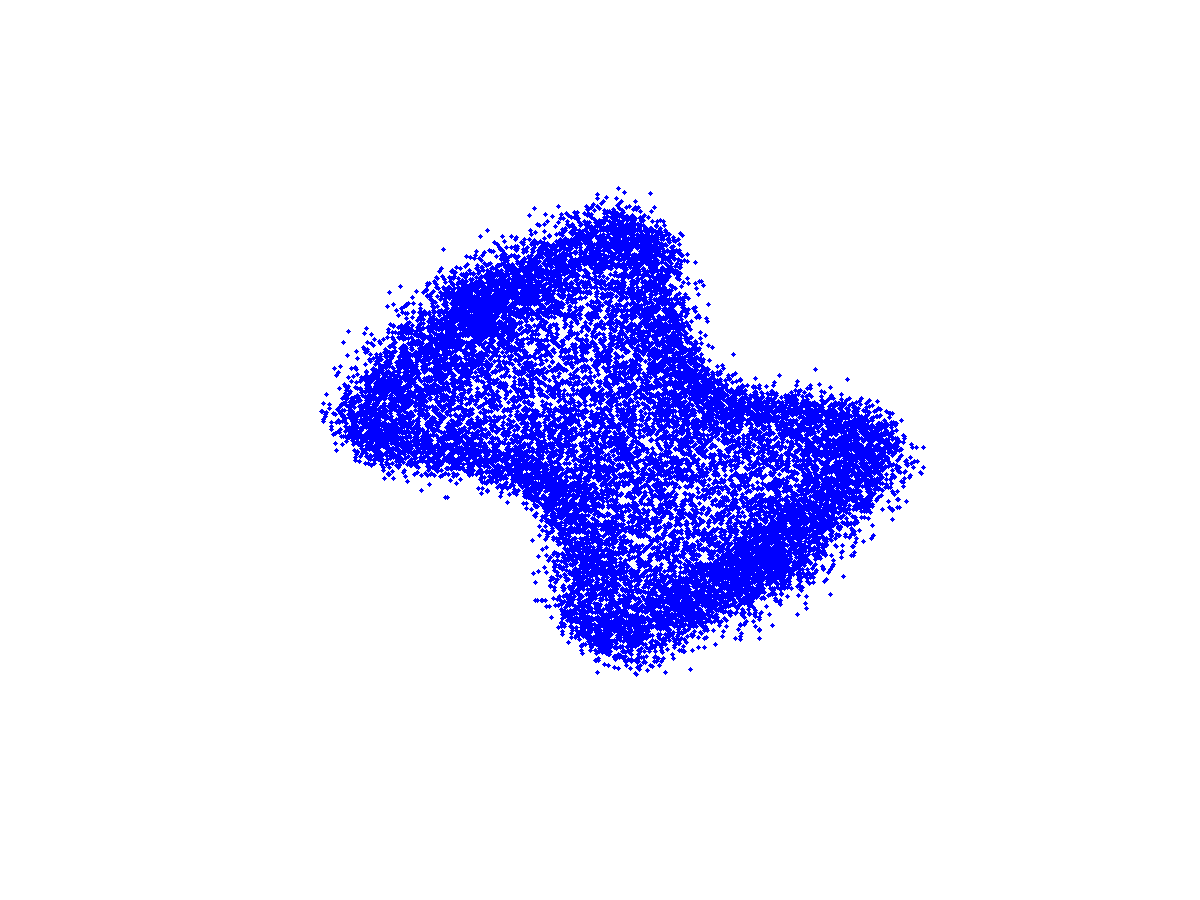}
&
\hspace{-2.3em}
\includegraphics[width=\picwi]{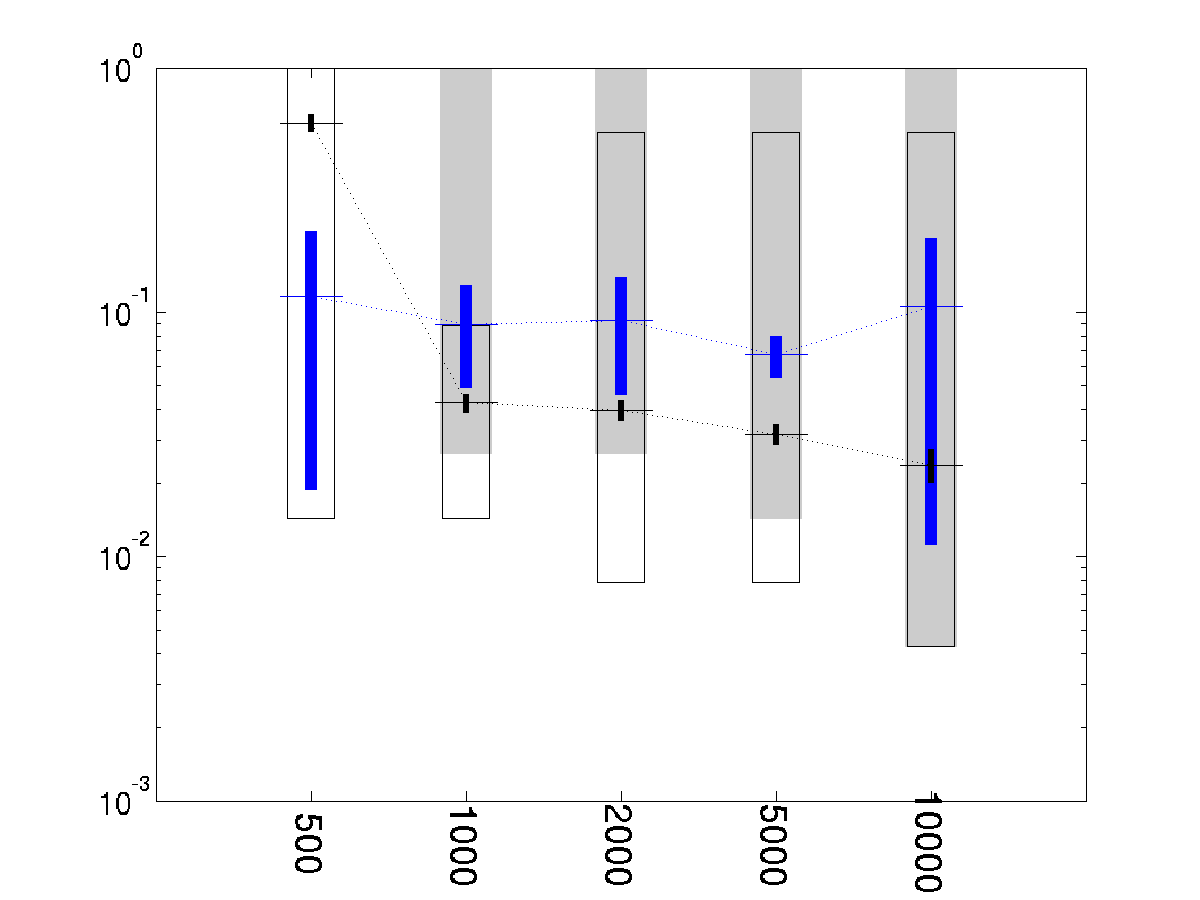}
&
\hspace{-2.3em}
\includegraphics[width=\picwi]{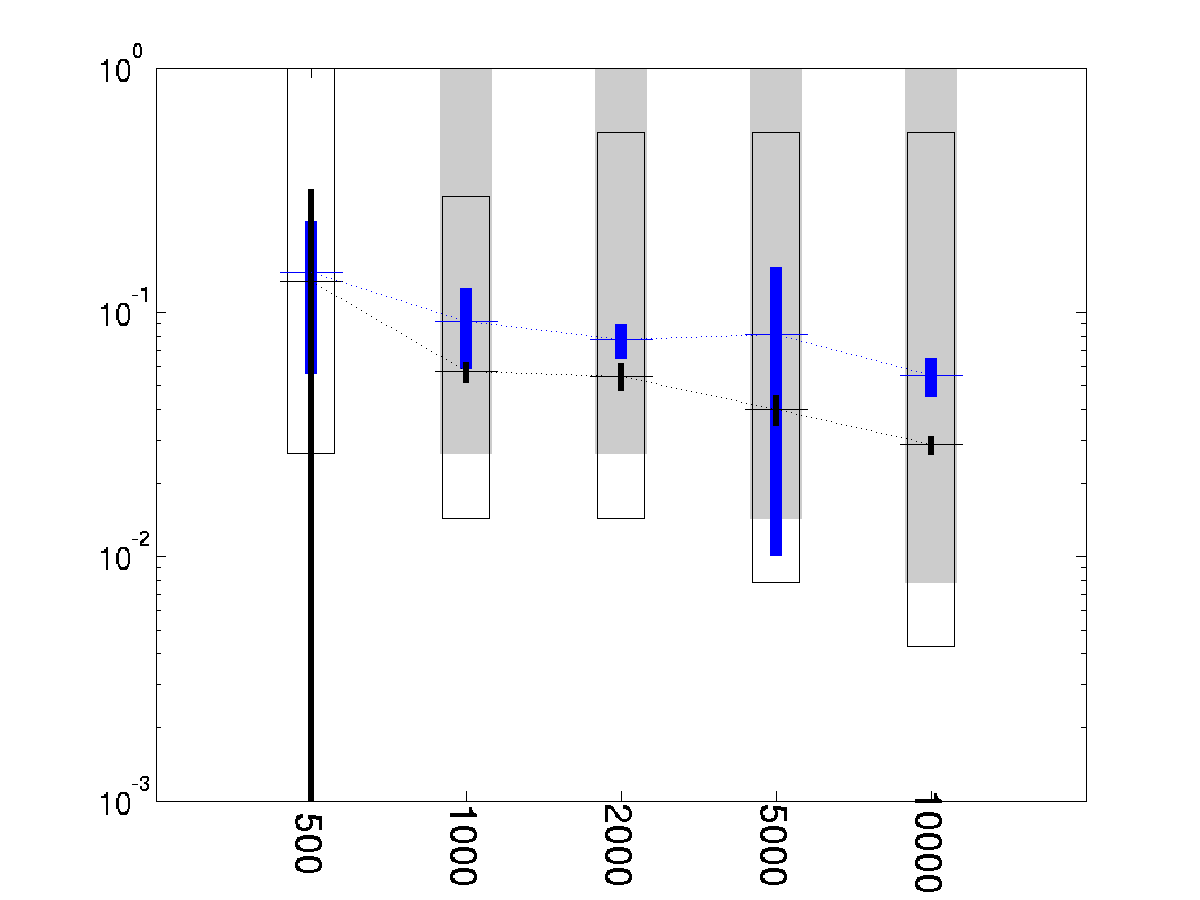}
&
\hspace{-2.3em}
\includegraphics[width=\picwi]{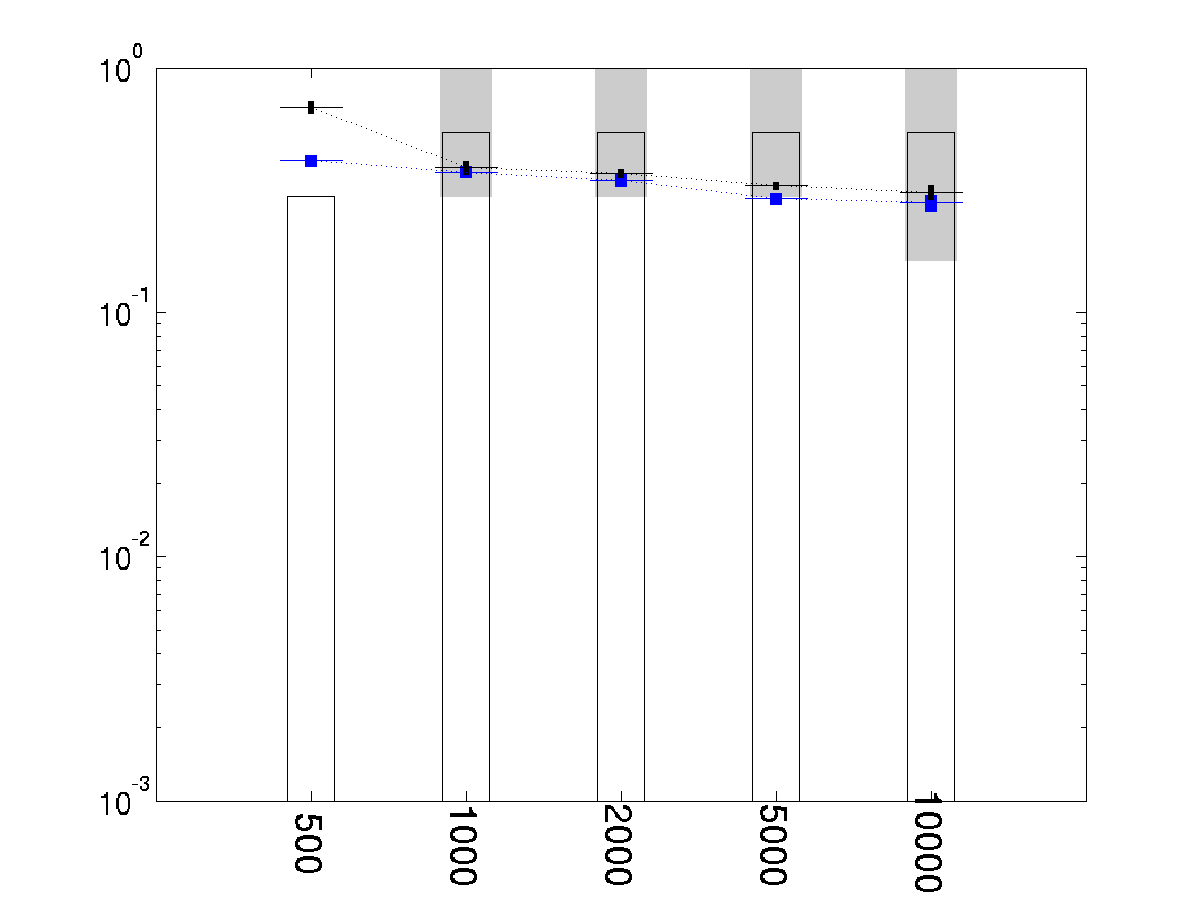}
\\
\end{tabular}
\caption{\small Estimates $\eppsh$ (mean and standard deviation over 10
  runs) on the {\tt dome} and {\tt hourglass} data, vs sample sizes
  $N$ for various noise levels $\sigma$; \textcolor{black}{$d'=2$} is
  in black and \textcolor{blue}{$d'=1$} in \textcolor{blue}{blue}. In
  the background, we also show as gray rectangles, for each $N,\sigma$
  the intervals in the $\epps$ range where the eigengaps of local SVD
  indicate the true dimension, and, as unfillled rectangles, the
  estimates proposed by
  \cite{Guangliang_ChenLittleMaggioniRosasco:multiscale11} for these
  intervals. \label{fig:synth-epps}}
\end{figure*}

{\bf Effects of $d'$, noise and $\nsamp$.}  The estimation results for $\epps$ are presented in Figure
\ref{fig:synth-epps}. As mentioned before, one could choose to
optimize the distortion in any number of dimensions $d'$ not exceeding
the intrinsic dimension $d$.  Let $\eppsh_{d'}$ denote the estimate
obtained from a $d'$ dimensional metric matching. We note a few
interesting things. First, when $d_1<d_2$, typically
$\eppsh_{d_1}>\eppsh_{d_2}$, but the values are of the same order (a
ratio of about 2 in the synthetic experiments). The explanation is
that, at $\epps$ values near the optimal one, chosing $d'<d$
directions in the tangent plane will select a subspace aligned
with the ``least curvature'' directions of the manifold, if any exist,
or with the ``least noise'' in the random sample. In these directions,
the data will tolerate more smoothing, which results in
larger $\eppsh$. The variance of $\eppsh$ observed is due to randomness
in the subsample $\nsamp'$ used to evaluate the distortion. \comment{The {\tt
hourglass} is a manifold with variable curvature, and the distortions
will differ depending on the curvature at the location of the $\nsamp'$
points.} The optimal $\epps$ decreases with $\nsamp$ and grows with the
noise levels, reflecting the balance it must find between variance and
bias. Note that for the {\tt hourglass} data, the highest noise level
of $\sigma=0.1$ is an extreme case, where the original manifold is
almost drowned in the 13-dimensional noise. Hence, $\epps$ is not only
commensurately larger, but also stable between the two dimensions and
runs. This reflects the fact that $\epps$ captures the noise
dimension, and its values are indeed just below the noise amplitude of
$0.1\sqrt{13}$.
The {\tt dome} data set exhibits the same properties discussed
previously, showing that our method is effective even for manifolds
with border.

{\bf Could $\eppsh$ be used to improve the estimation of the intrinsic
dimension $d$ by the CLMR \cite{Guangliang_ChenLittleMaggioniRosasco:multiscale11} method?} The CLMR method of
estimating the intrinsic dimension $d$ has two components: first, it
performs local SVD around each data point at a variety of scales
$\epps$ (this is akin to our weighted tangent plane
projections); then, it finds a range of scales in $\epps$ space, which
we shall call the CLMR range, where the largest eigengap is the $d$-th
eigengap. The $d$ is estimated by finding the largest eigengap
somewhere in the CLMR range. 

\begin{figure}
\setlength{\picwi}{0.38\textwidth}
\begin{center}
\begin{tabular}{cc}
\includegraphics[width=\picwi]{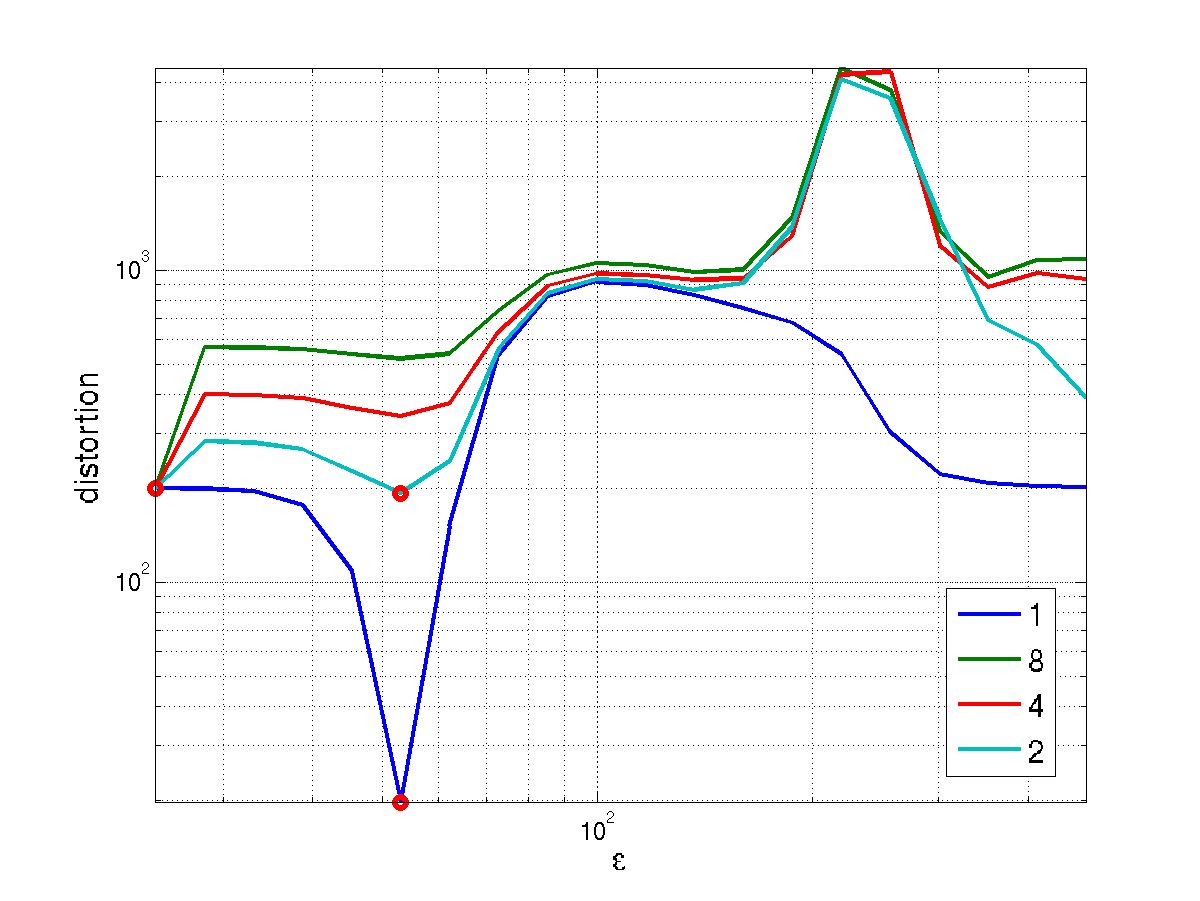} &
\includegraphics[width=\picwi]{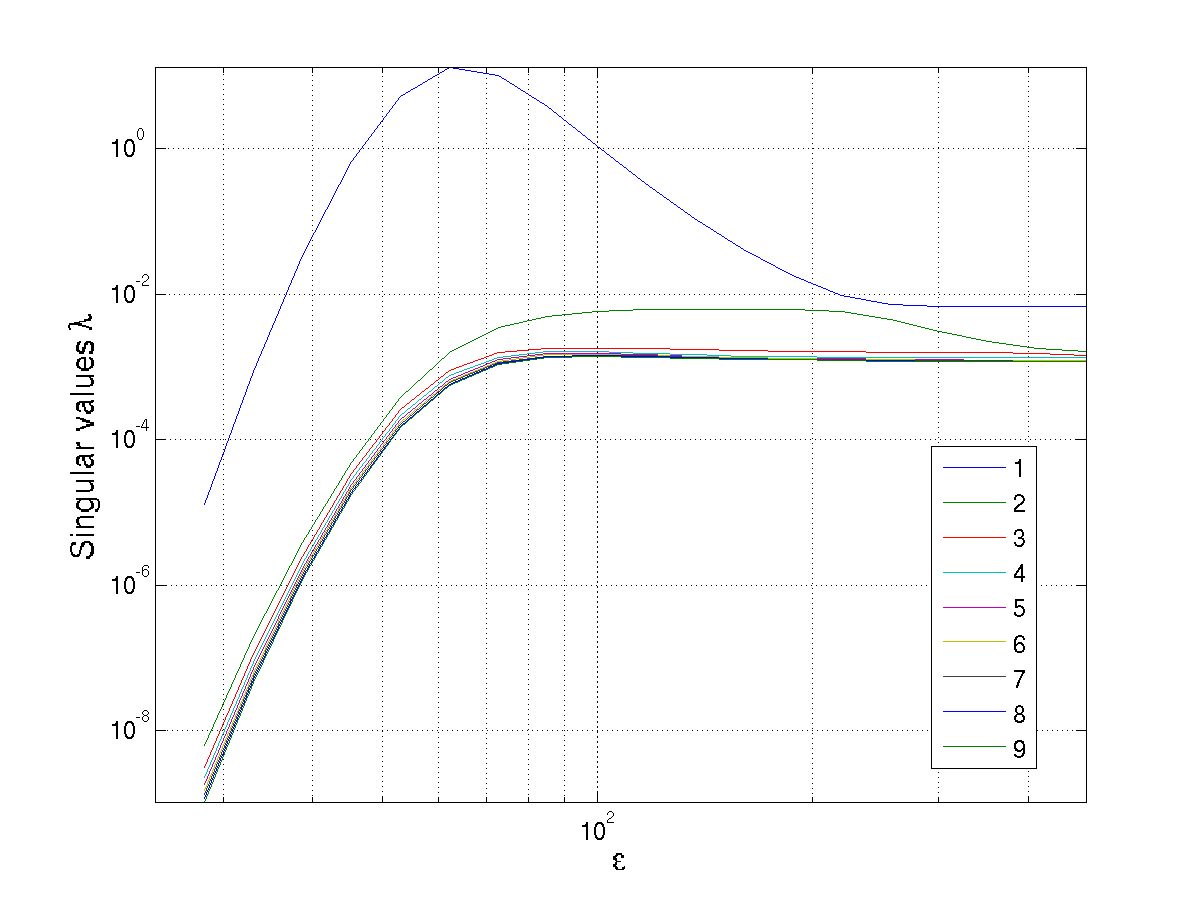} \\
\end{tabular}
\end{center}
\caption{\label{fig:ssl} 
{\tt COIL} data set (a) costs $\hat{\dis}$ for one sample of
$\nsamp'=200$ and $d'= 1,\,2,\,4\,8$, showing pronounced minimum at
$\eppsh=53.1$ for $d'=1$ (the lowest curve) and a weaker minimum for
$d'=2$; the range of $\epps$ searched was $[24, \,482]$ (b) the nine
largest singular values of local SVD versus $\epps$. We do not know
the intrinsic dimension of these high-dimensional data. The figure
shows why using a low dimensional projection, e.g. $d'=1$ may be a
practical strategy. One sees also that chosing $\epps$ by the CLMR
will result in values of at least $100-300$, depending which parameter
$K>1$ is chosen. The value chosen by crossvalidation is 54.}
\end{figure}

We computed the CLMR ranges both using the method
of \cite{Guangliang_ChenLittleMaggioniRosasco:multiscale11} (unfilled
rectangles in Figure \ref{fig:synth-epps}) and the ground truth ranges
(grey rectangles). As can be seen, our $\eppsh$ estimate {\em always}
lie in within the true ranges, meaning that if we computed the
eigengaps at $\eppsh$, we would find the true $d$, provided that such
a range exists. See \cite{Guangliang_ChenLittleMaggioniRosasco:multiscale11}
for a more detailed discussion of the limitations of this method in
e.g. high-dimensional noise. In contrast, the CLMR ranges only
partially overlap with the true ranges. We also found that the CLMR
method, which is based on finding the ``first descents'' of the
singular values, can be unreliable in that it may not find an upper or
a lower limit to the interval. Figure \ref{fig:ssl} illustrates this
phenomenon for the {\tt } data set used in the semi-supervised
experiments described below.  Note that the CLMR method depends on a
parameter $K$ to be set by the user, and we gave it the optimal $K$
for these data.
Figure \ref{fig:ssl} (a) shows the distortion $\dis$ that our
algorithm minimizes to find the optimal $\epps$ for the given data
set. Figure \ref{fig:ssl} (b) illustrates the range of $\epps$ chosen
by the CLMR method. The CLMR range is $[\epps_1, \epps_2]$ with
$\epps_1$ the smallest $\epps$ value for which $\lambda_{K+1}$ is
non-increasing and $\epps_2$ the smallest value for which $\lambda_1$
is non-decreasing. For this particular data set, the CLMR range is
approximately $[100,\,300]$ for $K>1$ ($K$ is an upper bound on the
intrinsic dimension $d$ of the data). Hence, the CLMR method would
choose an $\eppsh$ of at least 100 (200 if the middle of the CLMR
interval is used). 

{\bf Experiments with Smoothing.} To investigate whether the $\eppsh$
values chosen by our algorithm were ``good'' values for manifold
learning in noise, we sampled $\nsamp$ points from the {\tt hourglass} with no noise added, and we formed the sample $X^*$. Then we added
13-dimensional noise of amplitude $\sigma$ as described above,
obtaining the data set $X$, where each point $i$ of $X$ is the noisy
version of point $i$ in $X^*$. We embedded $X$ and $X^*$ into 3 dimensions using the same method (Laplacian Eigenmaps), obtaining coordinates $\phi_{\epps,i}$ and $\phi^*_{\epps^*,i}$, respectively, for
each point $i$. We aligned the two embeddings by the Procrustes method
and calculated the RMS error $\delta_{\epps,\epps^*}$ and
$\delta_{\epps}=\min_{\epps^*}\delta_{\epps,\epps^*}$. 
\begin{figure*}
\setlength{\picwi}{0.22\textwidth}
\begin{center}
\begin{tabular}{cccc}
\includegraphics[width=\picwi]{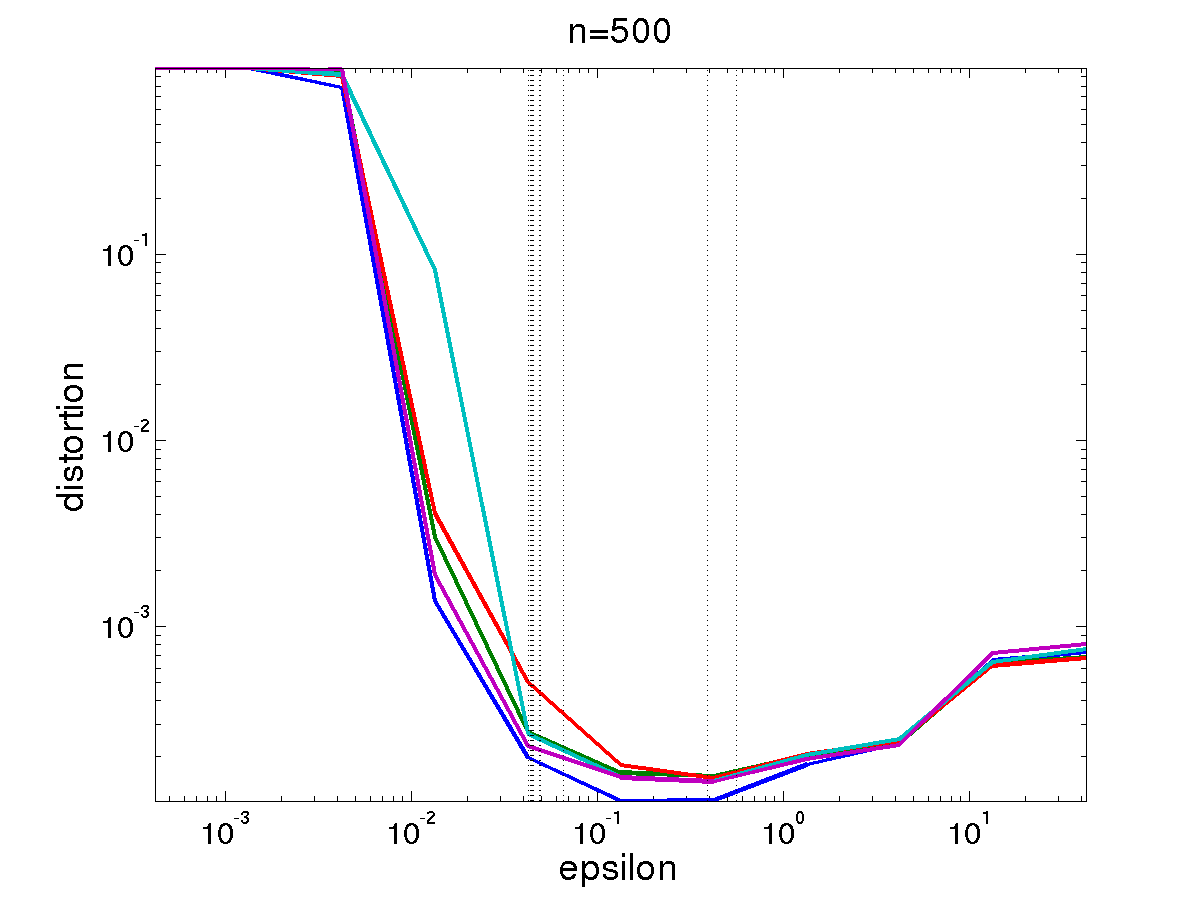} &
\includegraphics[width=\picwi]{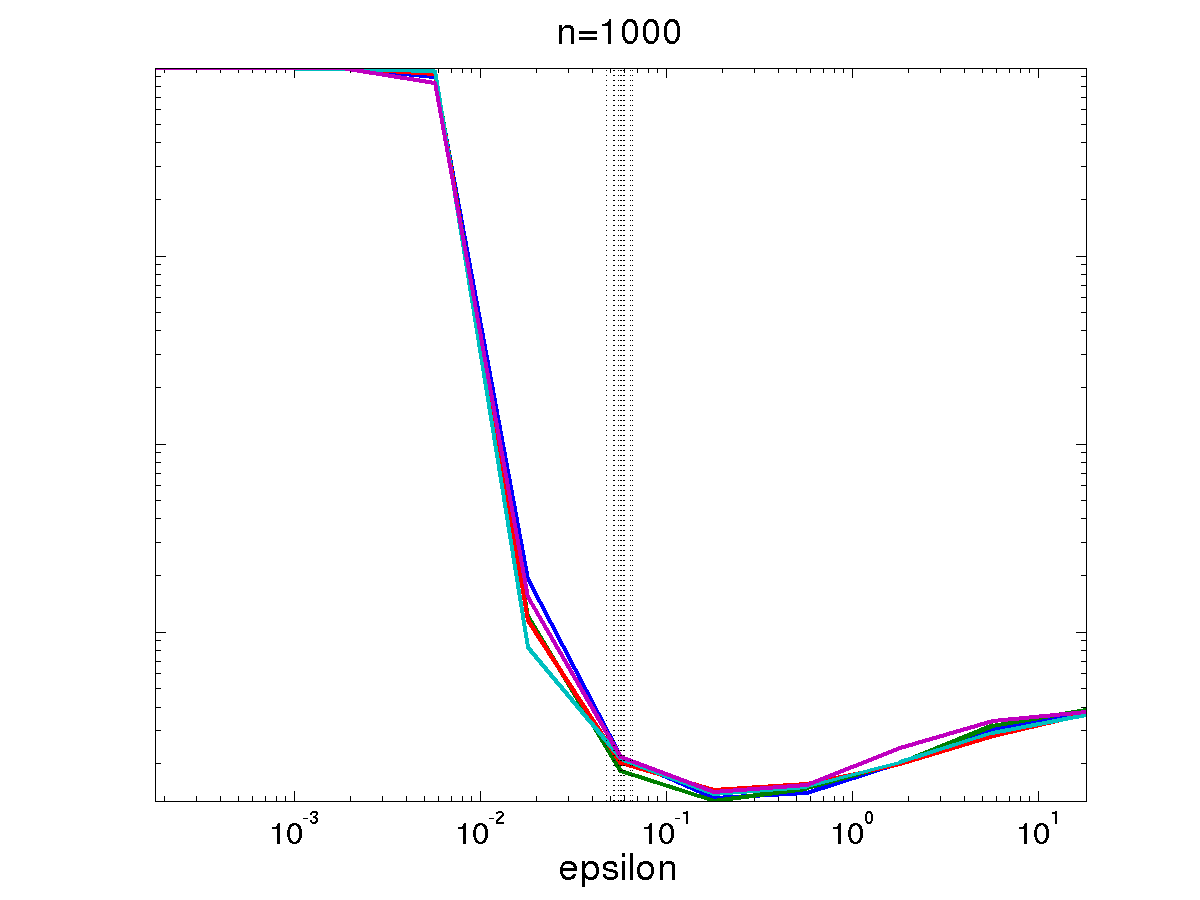} &
\includegraphics[width=\picwi]{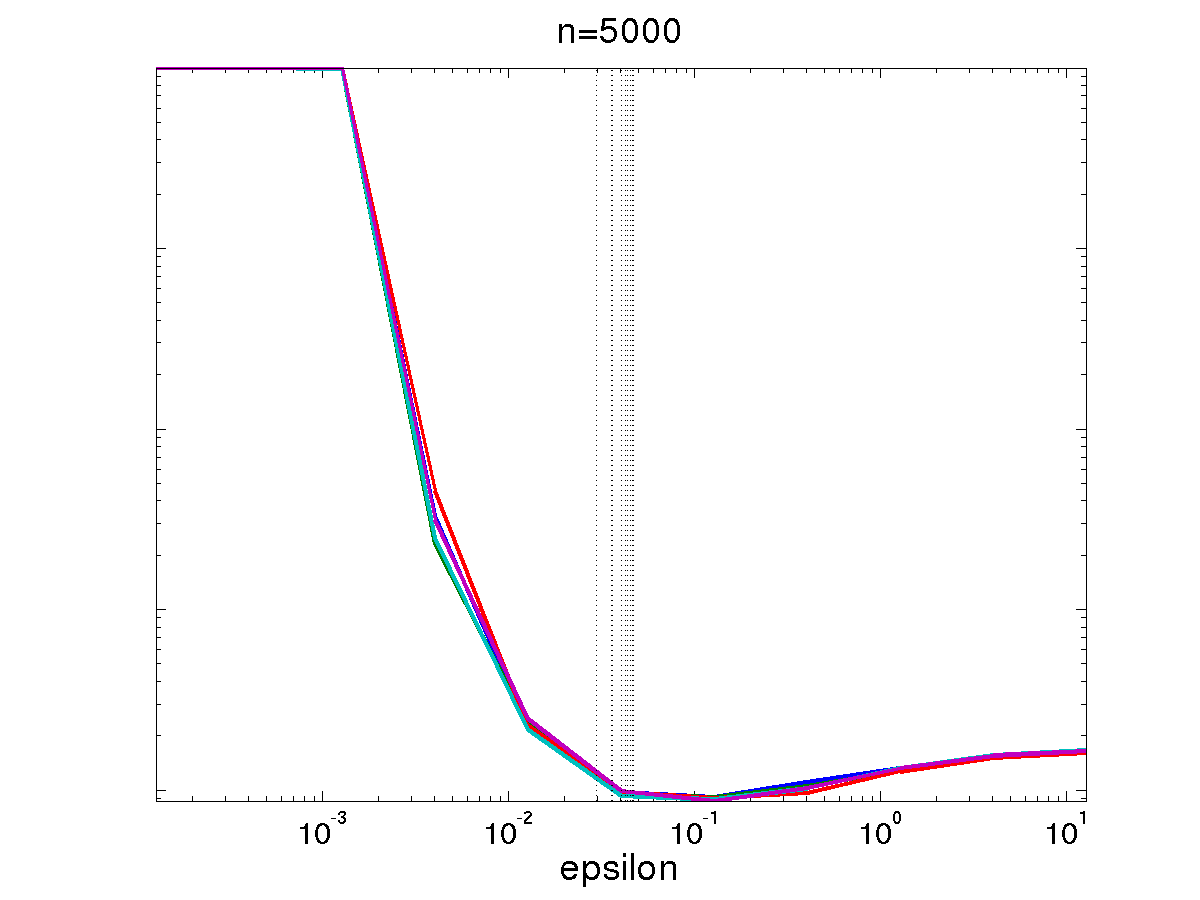} &
\includegraphics[width=\picwi]{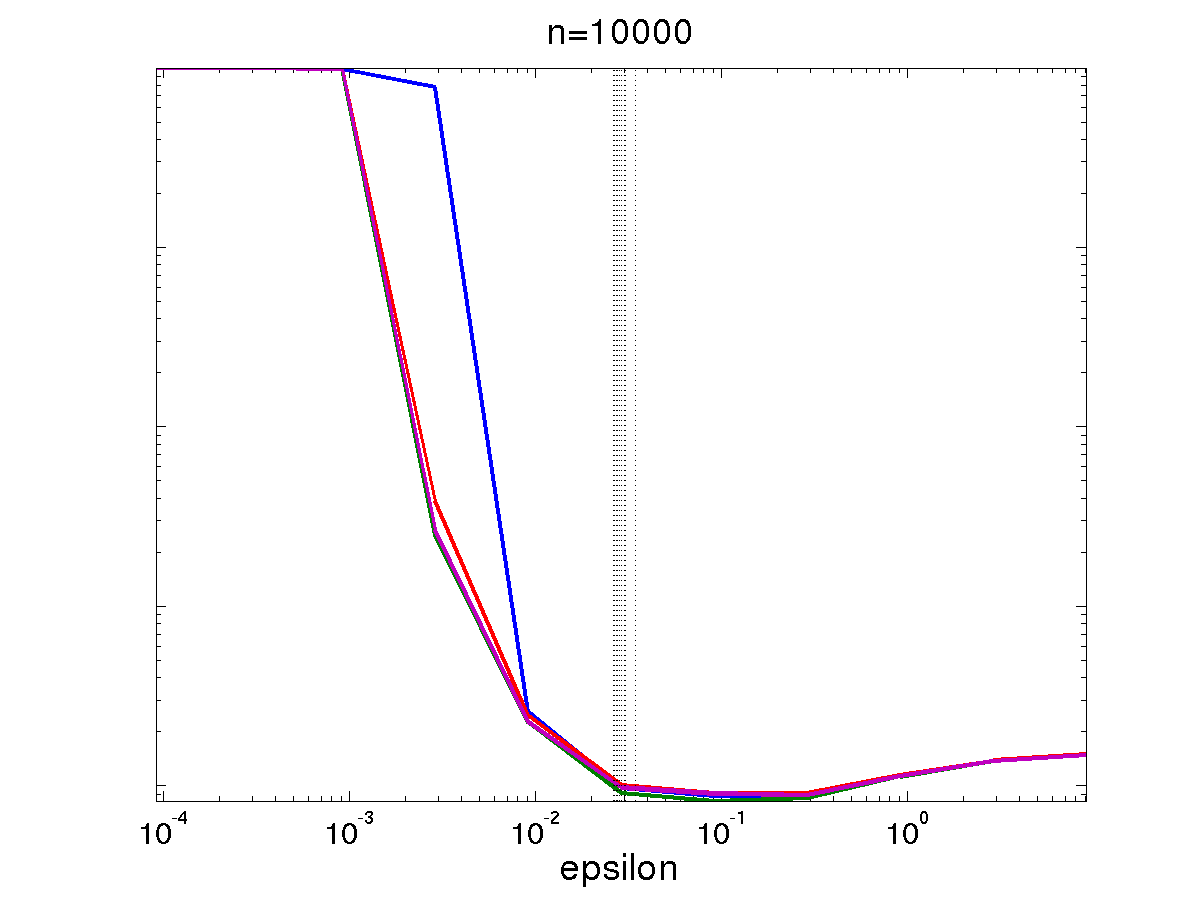}\\
\end{tabular}
\end{center}
\caption{\small \label{fig:smooth}
 Distortions between embedding of noisy and noiseless manifold data,
 for various $\epps$ values and sample sizes $\nsamp$. The manifold is the
 {\tt hourglass} embedded in 3D by Laplacian Eigenmap, data in 13
 dimensions, noise with $\sigma=0.001$; $\epps$ is the scale for the
 noisy data embedding, and the distortion shown is the lowest over all
 $\epps^*$ values for the noiseless data embedding; there were 5
 replications in each experiment. The vertical lines are the same
 $\hat{\epps}$ from Figure \ref{fig:synth-epps} (10
 replicates). 
}
\end{figure*}
In Figure \ref{fig:smooth}, 
 we show $\delta_\epps$ vs. $\epps$ (as ground truth)
along with $\eppsh$ obtained by our method (\selfco~). Our $\eppsh$ always finds a region of low $\delta_\epps$,
with a slight but systematic tendency to undershoot. Thus, the
experiment supports the case for choosing $\eppsh$ by
\eqref{eq:argmin_d} in unsupervised manifold learning, even when noise
is present (for which there is yet no theory).

{\bf Semi-supervised Learning (SSL) with Real Data.}  In this set of
experiments, the task is classification on the benchmark SSL data sets
proposed by \cite{ChapScho06}. This was done by least-square
classification, similarly to \cite{ZhouBelkin11}, after choosing the
optimal bandwidth by one of the methods below.
\bit
\item[{\tt{TE}}] \label{it:testerror}\textit{Minimize Test Error},
i.e. ``cheat'' in an attempt to get an estimate of the ``ground truth''.
\item[\tt{CV}] {\em Cross-validation} We split the training set
(consisting of 100 points in all data sets) into two equal
groups;\footnote{In other words, we do 2-fold CV. We also tried 20-fold and 5-fold CV, with no significant difference.} we use
simulated annealing to minimize the highly non-smooth cross-validation
classification error.
\item[{\tt{Rec}}] \label{it:eppslwPCA} \textit{Minimize the
reconstruction error} We cannot use the method of 
\cite{ChenBuja:localMDS09} directly, as it requires an embedding, so 
\comment{
A key characteristic of the manifold hypothesis is that the data is
locally linear, and the scale at which the data is locally linear is
that at which most manifold learning is performed
(\cite{AsBiTo11,Guangliang_ChenLittleMaggioniRosasco:multiscale11,Lee07NDR}). A
simple proxy can be used to find this scale without resorting to
estimating the intrinsic dimension. Indeed, one can simply use the
scale at which the data points are a weighted linear combination of
their neighbors. In our case, since we are constructing the Laplacian
from the heat kernel, we use that heat kernel as the basis for the
weights. Specifically,}we minimize reconstruction error based on the
heat kernel weights w.r.t. $\epps$ (this is reminiscent of LLE
\cite{SauRow03}):
$\mathcal{R}(\epps) = \sum_{i=1}^{n} \left | \left | \mbx_i - \sum_{j \neq i} \frac{w_{\epps}(\mbx_i,\mbx_j)}{\sum_{l \neq i}w_{\epps}(\mbx_i,\mbx_l)}\mbx_j \right | \right |^2 $

\comment{allowing us to report confidence regions for the
``optimal'' $\epps$.--I don't think that's a useful confidence region,
since it will be centered on an overfitted estimate. If we find a
meaning for the mean of these eps' we can change this text.}
\eit
Our method is denoted \selfco~ for {\em Geometric Consistency}; we
evaluate straighforward \selfco, that uses the dual Riemannian metric,
and a variant that includes the matrix inversion in Algorithm
\ref{alg:rmetric} denoted
\selfcomin.
\begin{table*}
\begin{small}
\begin{center}
\begin{tabular}{|c||c|c|c||c|c|}
\cline{2-6} 
\multicolumn{1}{c|}{} & {\tt TE} & {\tt CV} & {\tt Rec} & \selfcomin & \selfco \tabularnewline
\hline 
\hline 
\multirow{2}{*}{{\tt Digit1}} & 0.67$\pm$0.08 & 0.80$\pm$0.45 & \multirow{2}{*}{0.64} & \multirow{2}{*}{0.74} & \multirow{2}{*}{0.74} \tabularnewline
 & $[0.57,0.78]$ & $[0.47,1.99]$ &  &  &   \tabularnewline
\hline 
\multirow{2}{*}{{\tt USPS}} & 1.24$\pm$0.15 & 1.25$\pm$0.86 & \multirow{2}{*}{1.68} & \multirow{2}{*}{2.42} & \multirow{2}{*}{1.10} \tabularnewline
 & $[1.04,1.59]$ & $[0.50,3.20]$ &  &  &  \tabularnewline
\hline 
\multirow{2}{*}{{\tt COIL}} & 49.79$\pm$6.61 & 69.65$\pm$31.16 & \multirow{2}{*}{78.37} & \multirow{2}{*}{216.95} & \multirow{2}{*}{116.38} \tabularnewline
 & $[42.82,60.36]$ & $[50.55,148.96]$ &  &  &   \tabularnewline
\hline 
\multirow{2}{*}{{\tt BCI}} & 3.4$\pm$3.1 & 3.2$\pm$2.5 & \multirow{2}{*}{3.31} & \multirow{2}{*}{3.19} & \multirow{2}{*}{5.61} \tabularnewline
 & $[1.2,8.9]$ & $[1.2,8.2]$ &  &  &   \tabularnewline
\hline 
\multirow{2}{*}{{\tt g241c}} & 8.3$\pm$ 2.5 & 8.8$\pm$3.3  & \multirow{2}{*}{3.79 } & \multirow{2}{*}{7.37} & \multirow{2}{*}{7.38} \tabularnewline
 & $[6.3,14.6]$ & $[4.4,14.9]$ &  &  &  \tabularnewline
\hline 
\multirow{2}{*}{{\tt g241d}} & 5.7$\pm$ 0.24 & 6.4$\pm$1.15 & \multirow{2}{*}{3.77} & \multirow{2}{*}{7.35} & \multirow{2}{*}{7.36} \tabularnewline
 & $[5.6,6.3]$ & $[4.3,8.2]$ &  &  &   \tabularnewline
\hline 
\end{tabular}
\end{center}
\end{small}

\caption{Estimates of $\epps$ by  methods presented for the
six SSL data sets used, as well as TE. For TE and CV, which depend on
the training/test splits, we report the average, its standard error,
and range (in brackets below) over the 12 splits.\label{table:bw_est}}
\end{table*}

\begin{table}[!htb]

    \begin{minipage}{.5\linewidth}

      \centering
\begin{tabular}{|c||c|c|c|c|}
\cline{2-5} 
\multicolumn{1}{c|}{} & {\tt CV} & {\tt Rec} & \selfcomin & \selfco  \tabularnewline
\hline 
\hline 
{\tt Digit1} & 3.32 & 2.16 & 2.11 & 2.11 \tabularnewline
\hline 
{\tt USPS} & 5.18 & 4.83 & 12.00 & 3.89 \tabularnewline
\hline 
{\tt COIL} & 7.02 & 8.03 & 16.31 & 8.81 \tabularnewline
\hline 
{\tt BCI} & 49.22 & 49.17 & 50.25 & 48.67 \tabularnewline
\hline 
{\tt g241c} & 13.31 & 23.93 & 12.77 & 12.77 \tabularnewline
\hline 
{\tt g241d} & 8.67 & 18.39 & 8.76 & 8.76 \tabularnewline
\hline 
\end{tabular}

    \end{minipage}%
    \begin{minipage}{.5\linewidth}
      \centering

\begin{tabular}{|c||c||c|c|c|}
\cline{3-5} 
\multicolumn{1}{c}{} & \multicolumn{1}{c|}{} & $d'$=1 & $d'$=2 & $d'$=3 \tabularnewline
\hline 
\hline 
\multirow{2}{*}{{\tt Digit1}} & \selfcomin & 0.74 & 0.29 & 0.30\tabularnewline
\cline{2-5} 
 & \selfco & 0.74 & 0.77 & 0.78 \tabularnewline
\hline 
\multirow{2}{*}{{\tt USPS}} & \selfcomin & 2.42 & 2.31 & 3.88 \tabularnewline
\cline{2-5} 
 & \selfco & 1.10 & 1.16 & 1.18 \tabularnewline
\hline      
\multirow{2}{*}{{\tt COIL}} & \selfcomin & 116  & 87.4 & 128 \tabularnewline
\cline{2-5} 
 & \selfco & 187 & 179 & 187 \tabularnewline
\hline        
\multirow{2}{*}{{\tt BCI}} & \selfcomin & 3.32 & 3.48 & 3.65 \tabularnewline
\cline{2-5} 
 & \selfco & 5.34 & 5.34 & 5.34 \tabularnewline
\hline        \
\multirow{2}{*}{{\tt g241c}} & \selfcomin & 7.38  & 7.38 & 7.38 \tabularnewline
\cline{2-5} 
 & \selfco & 7.38  & 9.83 & 9.37 \tabularnewline
\hline        
\multirow{2}{*}{{\tt g241d}} & \selfcomin & 7.35 & 7.35 & 7.35 \tabularnewline
\cline{2-5}         
 & \selfco & 7.35 & 9.33 & 9.78 \tabularnewline
\hline 
\end{tabular}
    \end{minipage} 
        \caption{{\it Left panel}: Percent classification error for the six SSL data sets using the four $\epps$ estimation methods described. {\it Right panel}: $\epps$ obtained for the six datasets using various $d'$ values with \selfco~
and \selfcomin~. $\hat{\epps}$ was computed for
$d$=5 for {\tt Digit1}, as it is known to have an
intrinsic dimension of 5, and found to be 1.162 with \selfco~ and 0.797 with \selfcomin~. \label{table:dim_results}} 
\end{table}

Across all methods and data sets, the estimate of the bandwidth that
was furthest away from the ``optimal'' value determined by {\tt TE}
led to the highest classification error, see left panel of Table \ref{table:dim_results}. This confirms that
performance in classification when using a Laplacian-based regularizer
is quite sensitive to the estimate of the bandwidth of the Laplacian
and lends legitimacy to our attempt at finding a better, more
principled method for doing so.

Across five of the six data sets\footnote{In the
{\tt COIL} data set, despite their variability, {\tt CV}
estimates still outperformed the \selfco-based methods. This is the only data set constructed from a
collection of manifolds - in this case, 24 one-dimensional image
rotations. As such, one would expect that there would be more than one
natural length scale. 
}, cross-validation did not perform as well as the
\selfco-based methods, and took 2 to 6 times longer to compute. Further, the {\tt CV} estimates of $\epps$ in
each of the 12 training sets within a data set were highly variable,
with standard errors often of the same order as the estimated values
themselves. This suggests that {\tt CV} tends to overfit rather than
find values that generalize well. \comment{suggestion: it may be
unfair to report the classification error that comes out of using the
AVERAGE of the 12 values, since the cross-validation exercise never
actually yielded that estimate of the bandwidth. --Yeah, shall we
repeat this part? by averaging the test errors I mean}

{\bf Effect of Dimension $d'$.} One of the inputs required for
computing the distortion of \eqref{eq:d-data} is $d$, the intrinsic dimension
of $\M$. In most cases, $d$ is not known, and we do
not offer a new method for estimating it. However, we examine how changing the dimension to $d'\leq
d$ alters our estimate of $\epps$ and report our findings in the right panel of Table \ref{table:dim_results}.

The right panel of table \ref{table:dim_results} shows that the $\hat{\epps}$ for different $d'$ values are close, even though we search
over a range of two orders of magnitude. Even
for {\tt g241c} and {\tt g241d}, which were constructed so as to not
satisfy the manifold hypothesis, our method does reasonably well at
estimating $\epps$. That is, our method finds the $\hat{\epps}$ for which
the Laplacian encodes the geometry of the data set irrespective of
whether or not that geometry is lower-dimensional.

Overall, we have found that using $d'=1$ is most stable, and that
adding more dimensions introduces more numerical problems: it becomes
more difficult to optimize the distortion as in \eqref{eq:argmin_d},
as the minimum becomes shallower. \comment{Figure \ref{fig:distortion_minima}
in the Supplement illustrates this point nicely with data set {\tt
Digit1}.} In our experience, this is due to the increase in variance
associated with adding more dimensions. Using one dimension probably
works well because the wlPCA selects the dimension that explains the
most variance and hence is the closest to linear over the scale
considered. Subsequently, the wlPCA moves to incrementally ``shorter''
or less linear dimensions, leading to more variance in the estimate of
the tangent plane.

\section{Discussion}

In manifold learning, supervised and unsupervised, estimating the
graph versions of Laplacian-type operators is a fundamental task. We
have provided a principled method for selecting the parameters of such
operators, and have applied it to the selection of the bandwidth/scale
parameter $\epps$. Moreover, our method can be used to optimize any other parameters used in the graph Laplacian; for example, $k$ in the
$k$-nearest neighbors graph\comment{\footnote{In conjunction with
\cite{TingHuangJordan10}}}, or - more interestingly - the renormalization parameter $\lambda$ \cite{CoiLaf06} of
the kernel. The latter is theoretically equal to 1, but it is possible that
it may differ from 1 in the finite $\nsamp$ regime. In general, for finite $\nsamp$, a
small departure from the asymptotic prescriptions may be beneficial - 
and a data-driven method such as ours can deliver this benefit.

By imposing geometric self-consistency, our method estimates an {\em
intrinsic quantity} of the data. \selfco~ is also fully unsupervised,
aiming to optimize a (lossy) representation of the data, rather than a
particular task. This is an efficiency if the data is used in an
unsupervised mode, or if it is used in many different subsequent
tasks. Of course, one cannot expect an unsupervised method to
always be superior to a task-dependent one. Yet, \selfco~ has shown to
be competitive and even superior in experiments with the widely
accepted {\tt CV}. Besides the experimental validation, there are
other reasons to consider an unsupervised method like \selfco~ in a
supervised task: (1) the labeled data is scarce, so $\eppsh$ will
have high variance, (2) the CV cost function is highly non-smooth
while $\dis$ is much smoother, and (3) when there is more than one
parameter to optimize, difficulties (1) and
(2) become much more severe. 

Our algorithm requires minimal prior knowledge. In particular, it {\em
does not} require exact knowledge of the intrinsic dimension $\dintri$,
since it can work satisfactorily with $d'=1$ in many cases.

\comment{The asymptotic consistency of $\dis$ remains for now an open problem, but the supplement offers some theoretical insights on the choice of norm in $\dis$ as well as in the soundness of using $d'<d$.} 

An interesting problem that is outside the scope of our paper is the
question of whether $\epps$ needs to vary over $\M$. This is a
question/challenge facing not just \selfco, but any method for setting
the scale, unsupervised or supervised. Asymptotically, a uniform $\epps$
is sufficient. Practically, however, we believe that allowing $\epps$ to vary may be
beneficial. In this respect, the \selfco~method, which simply
evaluates the overall result, can be seamlessly adapted to work with
any user-selected spatially-variable $\epps$, by appropriately
changing \eqref{eq:heat_kernel} or sub-sampling $\dataset$ when
calculating $\dis$.


\subsection*{Acknowledgments} 
This work was partially supported by awards IIS-0313339 and
EEC-1028725 from NSF.The content is solely the responsibility of the
authors and does not necessarily represent the official views of the
National Science Foundation. The authors also gratefully acknowledge
NSF award IIS-0313339 under which ideas for this research originated.


\bibliography{thesis}
\bibliographystyle{apalike}

\end{document}